\documentclass[pdflatex,sn-mathphys-num]{sn-jnl}

\usepackage{graphicx}%
\usepackage{booktabs, tabularx}%
\usepackage{multirow}%
\usepackage{amsmath,amssymb,amsfonts}%
\usepackage{amsthm}%
\usepackage{mathrsfs}%
\usepackage[title]{appendix}%
\usepackage{xcolor}%
\usepackage{textcomp}%
\usepackage{manyfoot}%
\usepackage{booktabs}%
\usepackage{algorithm}%
\usepackage{algorithmicx}%
\usepackage{algpseudocode}%
\usepackage{listings}%

\theoremstyle{thmstyleone}%
\theoremstyle{thmstyletwo}%

\theoremstyle{thmstylethree}%

\begin{document}

\title[Article Title]{Comparative Analysis of NMPC and Fuzzy PID Controllers for Trajectory Tracking in Omni-Drive Robots: Design, Simulation, and Performance Evaluation}


\author{\fnm{Love} \sur{Panta}}\email{075bei016.love@pcampus.edu.np}

\affil{\orgdiv{Electronics and Computer Engineering}, \orgname{Pulchowk Campus}, \orgaddress{\city{Lalitpur}, \postcode{44600}, \state{Bagmati}, \country{Nepal}}}


\abstract{Trajectory tracking for an Omni-drive robot presents a challenging task that demands an efficient controller design. This paper introduces a self-optimizing controller, Type-1 fuzzyPID, which leverages dynamic and static system response analysis to overcome the limitations of manual tuning. To account for system uncertainties, an Interval Type-2 fuzzyPID controller is also developed. Both controllers are designed using Matlab/Simulink and tested through trajectory tracking simulations in the CoppeliaSim environment. Additionally, a non-linear model predictive controller (NMPC) is proposed and compared against the fuzzyPID controllers. The impact of tunable parameters on NMPC's tracking accuracy is thoroughly examined. We also present plots of the step-response characteristics and noise rejection experiments for each controller. Simulation results validate the precision and effectiveness of NMPC over fuzzyPID controllers while trading computational complexity. Access to code and simulation environment is available in the following link: \href{https://github.com/love481/Omni-drive-robot-Simulation.git}{https://github.com/love481/Omni-drive-robot-Simulation.git}.}




\maketitle

\section{Introduction}
In recent years, the Omni-directional mobile robot has garnered attention and interest from various research communities due to its unique drive mode, robust maneuver control, and diverse applications across different fields. In comparison to various driving techniques, Omni-based robots offer controlled motion in all directions, allowing them to track user-defined paths or optimal trajectories with low computation costs \cite{shabalina2018comparative, taheri2020omnidirectional}.

The development of an optimal path for a robot to navigate in both static and dynamic environments has been a persistent concern for researchers. Several path-generating algorithms, such as Grassfire, Dijkstra, $A^*$, and RRT, have been extensively used in mobile robotics to provide optimal solutions \cite{karur2021survey}. Among these algorithms, $A^*$ is commonly employed for 2D plane navigation, utilizing a heuristic approach to generate the desired trajectory. However, the paths generated by $A^*$ often consist of straight segments and sharp angular turns, which may be undesirable for navigating robots. To address this, path-smoothing techniques are applied to soften breakpoints and enhance parametric continuity \cite{ravankar2018path}.

Conventional PID controllers \cite{cervantes2001pid} have been widely employed in various robotics application for efficient speed control \cite{somwanshi2019comparison} and waypoint tracking \cite{lee2018design}. However, optimizing these controllers for desirable output characteristics can be challenging, and they may not be suitable for a wide range of operating conditions \cite{somwanshi2019comparison, jiang2012design}. Consequently, an alternative and effective approach to controller design is utilizing fuzzy logic. Fuzzy logic controllers leverage sets of rules derived from the responses of various static and dynamic systems to adjust tuning parameters based on fuzzy input variables \cite{bansal2013speed, ghanim2020optimal, chao2019optimal, mendel2017uncertain}. These controllers exhibit good system performance, transient response, and disturbance rejection capabilities. Moreover, they have been successfully applied to various non-linear plants that are challenging to model due to a lack of sufficient parameters \cite{nour2007fuzzy, alouache2018fuzzy}. 

In the context of autonomous path tracking, a fuzzy-based controller has been proposed that leverages the geometric properties of the path ahead for a differential drive robot \cite{antonelli2007fuzzy}. This controller considers the curvature and distance to the next bend relative to the robot's current location, thereby mimicking real driving behavior to determine the desired cruise speed. Similarly, T1-Fuzzy PID controllers have been developed using 25 fuzzy rules for the automatic tuning of PID gains in differential drive robots and quadcopters \cite{lee2018design, rabah2018design}. The tuned gains are then used to compute the desired velocity, guiding the robot along the reference trajectory and minimizing tracking errors, which demonstrated superior performance compared to classical PID controllers. A Particle Swarm Optimization (PSO)-optimized fuzzy controller has been introduced that directly outputs joint torques for a 2-DOF planar robot \cite{bingul2011fuzzy}. In this approach, PSO is employed to optimize the antecedent and consequent parameters of the Mamdani-based fuzzy logic system. For omni-drive robots, studies have successfully evaluated the performance of fuzzy logic-based path planners \cite{hashemi2011model, abiyev2017fuzzy, masmoudi2016fuzzy}. These approaches continuously update control signals derived from the fuzzy logic controller based on deviation errors, guiding the robot to the desired waypoints.

Recently, significant research has focused on developing effective Type-2 Fuzzy Logic Controllers (FLCs) to model various sources of uncertainties in nonlinear systems \cite{mendel2017uncertain, mittal2020comprehensive, kumbasar2014simple}. A Type-2 Fuzzy Logic Controller has been presented for modeling uncertainties in measurements during the tracking of mobile objects in robotic soccer games \cite{figueroa2005type}. This controller uses angle error and change in angle as inputs to the fuzzy system, outputting speeds and directions to control the motion of the agents. Similarly, a novel application of GA-optimized IT2-FPD+I controllers has been explored for 5-DOF redundant robot manipulators to track desired joint trajectories \cite{kumar2017evolving}. Their experiments demonstrated robustness in terms of disturbance rejection, model uncertainties, and noise injections, outperforming previous T1-FPD and conventional PID controllers. A new method for building an optimal structure of a PID-type IT2 FLC system has been proposed for controlling joint actions in delta robots \cite{lu2016optimal}. This approach uses the Non-dominated Sorting Genetic Algorithm (NSGA) to tune the controller's scaling factors by formulating the problem as a multi-objective optimization task. In another study, Social Spider Optimization (SSO) has been utilized for parameter tuning \cite{humaidi2021social}. Additionally, the effects of Footprint of Uncertainty (FOU) parameters on generating the control surface for a Single Input IT2-Fuzzy PID (SI-IT2-FPID) controller have been explained, with performance demonstrations for trajectory tracking of UAVs \cite{sarabakha2017type}.

With the advancement of powerful computing devices, optimization-based control techniques have become integral in various autonomous mobile industries \cite{vu2021model}. Model Predictive Control (MPC) is one such method widely used for path tracking of mobile robots, as it takes into consideration various constraints for optimal control inputs \cite{maurovic2011explicit, liu2022mpc, kanjanawanishkul2009path, yang2018trajectory, nascimento2018nonlinear}. In comparison with traditional PID methods, MPC offers higher accuracy, smoother control inputs, and increased resistance to external disturbances. Papers such as \cite{pacheco2015testing, liu2022mpc} also highlight the superior performance of MPC over PID controllers, especially when considering kinematic constraints in mobile robots. Considering this, different variants of MPC-based path tracking controllers have been evaluated in recent years, based on the nature of the prediction model's nonlinearity and the representation of output state quantities \cite{mayne2014model, bai2019review}. In 2007, the paper by Conceiccao \textit{et al.} \cite{conceiccao2007nonlinear} proposed non-linear MPC based on error kinematics models for Omni-drive robots for path following, which was particularly challenging due to hardware constraints at that time. To address these challenges, linear MPC was introduced, which linearized the robot model to simplify computation operations. In a subsequent paper by Wang \textit{et al.} \cite{wang2018trajectory}, the authors designed linear MPC and demonstrated its effectiveness for point stabilization and trajectory tracking. Similarly, Kanjanawanishkul \textit{et al.} \cite{kanjanawanishkul2009path} presented and applied linear error MPC in real-time for Omni-drive robots, considering system and input constraints to trade for stability. Furthermore, Gr{\"u}ne \textit{et al.} \cite{grune2011nonlinear} introduced various extensions of NMPC for time-varying references with or without terminal constraints. Despite advancements in trajectory tracking, few approaches have been applied to Omni-directional robots, leaving a gap in proposing Nonlinear Model Predictive Control (NMPC) and assessing its performance against other control methods.

We utilize CoppeliaSim as the simulation platform to conduct our experiments. We design the physical model of a four-wheel omni-drive robot in SolidWorks and import the URDF format into CoppeliaSim. Subsequently, it is interfaced with Matlab/Simulink via an external API to enable control. Physical parameters such as frictional force and air resistance are maintained at default values in the simulation environment. In summary, our paper comprises three main sections. The initial two sections are dedicated to modeling the kinematics of the robot and subsequently addressing path planning. Following this, we introduce an Omni-drive controller, employing two distinct methods: fuzzy logic with the Mamdani model and Non-linear Model Predictive Control (NMPC). We proceed to assess and compare the performance of these approaches in guiding the robot along the reference trajectory generated by the path planning algorithms. The design of the Omni-drive controller involves formulating a non-linear predictive model, treated as a cost minimization problem. The objective is to determine optimal control inputs that adhere to bounded input constraints, subsequently applied to the robot kinematics. This ensures the robot effectively follows the desired path outlined by the path planning algorithms.

\section{Kinematics model}
To derive the kinematics model of an Omni-drive robot, it is essential to be aware of the geometric configuration of each wheel with respect to the robot's Center of Gravity (COG). This knowledge is crucial for understanding how the global velocity of the robot is distributed to each wheel, assuming the absence of roller skidding. The kinematics model establishes the relationship between wheel angular velocity and the robot's global velocity, and vice versa. In this context, the global coordinate are aligned with the local coordinate frame of the robot. We extend the approach taken by \cite{baede2006motion} to derive the kinematic of 4-wheel Omni-drive robot.
\par
\begin{figure}
\begin{minipage}[c]{0.45\textwidth}
  \includegraphics[width=1.1\linewidth]{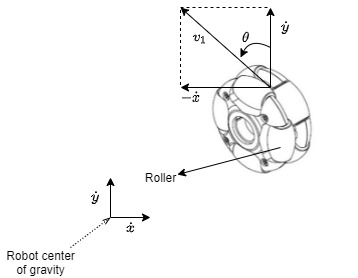}
  \caption{Component division of translational velocity of omni wheel}\label{omni_wheel}
\end{minipage}
\hspace{4 mm}
\begin{minipage}[c]{0.45\textwidth}
  \includegraphics[width=1.02\linewidth]{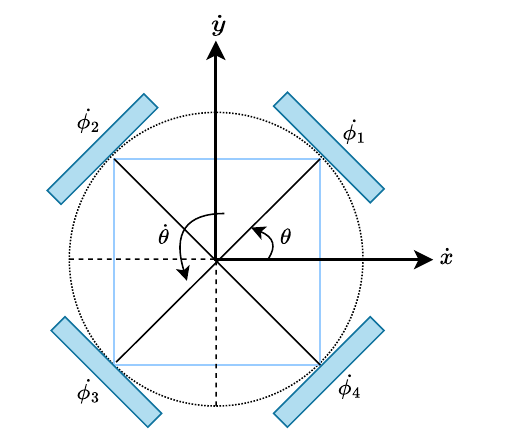}
  \caption{Kinematic diagram of four-wheel omni robot}\label{omni_model}
\end{minipage}%
\end{figure}

The global velocity of robot is written as ($\dot x$, $\dot y$, $\dot \theta$) and angular velocity of each wheel is denoted by ($\dot \phi_1$, $\dot \phi_2$, $\dot \phi_3$, $\dot \phi_4$) which are at an angle of $\dfrac{\pi}{4}$, $\dfrac{3\pi}{4}$, $\dfrac{5\pi}{4}$, $\dfrac{7\pi}{4}$ respectively. Robot body radius and wheel radius are also taken as $R$ and $r$. Then the translation velocity of each wheel hub $v_i$ can be formed as the combination of pure translation and pure rotational part of robot \cite{baede2006motion}.
Thus, Fig. \ref{omni_wheel} shows each wheel has a velocity equal to the expression given as,
\begin{equation}
v_i = -\sin(\theta + \alpha_i)\dot x + \cos(\theta + \alpha_i)\dot y + R\dot \theta
\label{kinematic_eq_for_wheel}
\end{equation}
where, $\theta$ + $\alpha_i$ is the offset angle of each wheel w.r.t COG. $\theta$ is $\dfrac{\pi}{4}$ in case of a four-wheel omni robot as shown in Fig. \ref{omni_model} and $\alpha_i$ is $\dfrac{\pi}{2}(i-1)$ for the wheel i=1,...,4 respectively. Now, the kinematics model of the robot can be represented in matrix form using equation \eqref{kinematic_eq_for_wheel} as,
\begin{gather}
\begin{bmatrix}\dot \phi_1 \\ \dot \phi_2 \\ \dot \phi_3 \\ \dot \phi_4 \end{bmatrix}
=
 \frac{1}{r}
\begin{bmatrix}-\sin(\theta + \alpha_1) & \cos(\theta + \alpha_1) & R \\
               -\sin(\theta + \alpha_2) & \cos(\theta + \alpha_2) & R \\
               -\sin(\theta + \alpha_3) & \cos(\theta + \alpha_3) & R \\
               -\sin(\theta + \alpha_4) & \cos(\theta + \alpha_4) & R \\
\end{bmatrix}
\begin{bmatrix}\dot x \\ \dot y \\ \dot \theta \end{bmatrix}
\end{gather}
The provided model is employed to address both forward and inverse kinematics problems essential for tracking the reference trajectory of the robot, as detailed in the subsequent sections of this paper.

\section{Path Planning}
We select the $A^*$ algorithm for its efficiency and its appropriateness in devising optimal paths for static environments \cite{karur2021survey}. This algorithm uses the heuristic approach to estimate the best optimal path by avoiding obstacles in the given 2D environment if the solution really exists. Here, the environment of the robot is discretized into a number of small grid cells called an occupancy grid map filled with the binary digit of '1' and '0' which denotes the obstacle and free space respectively. In Coppeliasim, we have created the custom static obstacles and build the map by treating robot as a point object. So, this increases the map area around the obstacles thereby easing the path planning problem for $A^*$ algorithm. In the following paper \cite{jing2018application}, different heuristic functions were compared that best optimize the algorithm. But, the simple way is to use Manhattan distance which is based upon the sum of the absolute differences between the two vectors i.e current node to goal node which is given as,
\begin{equation}
\label{manhattan distance}
ManhattanDistance(c) = |x_c-x_g| + |y_c-y_g|
\end{equation}
where subscript 'c' is taken as the current node and 'g' is the goal node.\\
So, the total cost function to be minimized for each node can be calculated as,
\begin{equation}
f(c) = g(c) + h(c)
\end{equation}
Here, $g(c)$ is the cost value from the starting node to the current node, and $h(c)$ is the heuristic Manhattan function from Equation \eqref{manhattan distance}. 
In order to smooth the path generated by $A^*$ algorithm, we choose to apply B-spline over bezier curve due to its superior ability to control local points \cite{ravankar2018path}. Moreover, it exhibits $C^2$ continuity which is important for stability and passenger comfort.

\section{Design of tracking Controller}
Given a smooth trajectory generated by B-spline, we obtain $N$ waypoints uniformly sampled at sampling rate of $t_s$ seconds. Here, $N$ is varying depending on the total time defined to reach the destination from the starting point. So, our trajectory waypoints of target robot can be formulated as,
\begin{equation}
    \mathbf{X}_{ref}(T)=[\mathbf{x}_{ref}(0), \mathbf{x}_{ref}(t_s),..., \mathbf{x}_{ref}(nt_s),..., \mathbf{x}_{ref}((N-1)t_s)]^T
\end{equation}
\begin{equation}
     \mathbf{U}_{ref}(T)=[ \mathbf{u}_{ref}(0),  \mathbf{u}_{ref}(t_s),...,  \mathbf{u}_{ref}(nt_s),..., \mathbf{u}_{ref}((N-1)t_s)]^T
\end{equation}
where, $T$ is the total specified tracking time, $\mathbf{x}_{ref}(nt_s)$ and $ \mathbf{u}_{ref}(nt_s)$ are the intermediate reference waypoint and corresponding velocity respectively. Each waypoint is represented as target robot pose coordinate by $[x_{ref}(nt_s), y_{ref}(nt_s), \theta_{ref}(nt_s)]$ at time $nt_s$. Here, reference heading angle $\theta_{ref}(nt_s)$ is calculated as,
\begin{equation}
    \label{eqn:heading error}
    \theta_{ref}(nt_s)= \tan^{-1}\left(\frac{\Delta y_{ref}(nt_s)}{ \Delta x_{ref}(nt_s)}\right)
\end{equation}

\begin{figure}[t!]
\includegraphics[width=0.7\linewidth]{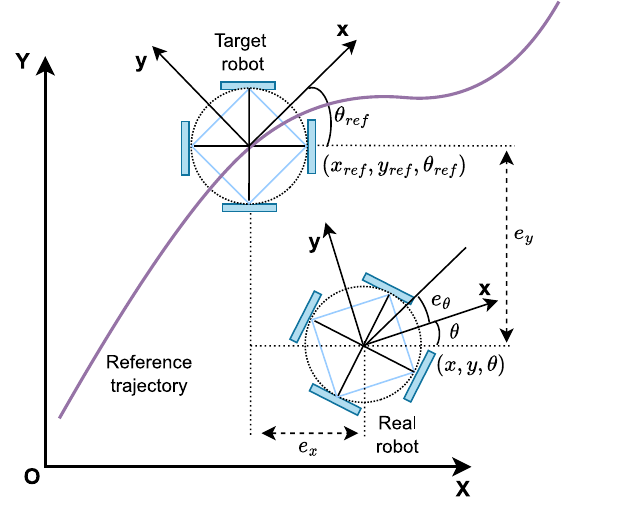}
\caption{Trajectory tracking error for Omni-drive robot}
\label{TTE}
\end{figure}
Similarly, we use $[v_{ref}(nt_s),\omega_{ref}(nt_s)]$ to represent reference linear and angular velocity at each sampling time which is given as,
\begin{equation}
   \begin{cases}
    v_{ref}(nt_s)= \frac{\sqrt{(\Delta x_{ref}(nt_s))^2+(\Delta y_{ref}(nt_s))^2}}{t_s} \\
   \omega_{ref}(nt_s)= \frac{\theta_{ref}(nt_s)-\theta_{ref}((n-1)t_s)}{t_s}
    \end{cases}
\end{equation}
such that, 
\begin{equation}
    \begin{cases}
    \Delta y_{ref}(nt_s)= y_{ref}(nt_s)-y_{ref}((n-1)t_s) \\
    \Delta x_{ref}(nt_s)= x_{ref}(nt_s)-x_{ref}((n-1)t_s)
     \end{cases}
\end{equation}
In order to track the generated waypoints, the robot position state must also be known in global coordinate system. It is denoted as $ \mathbf{X}(T)$ which compose of sequence of pose coordinates of real robot calculated at each sampling interval given by $[x(nt_s), y(nt_s), \theta(nt_s)]$. The whole process can be seen in Fig. \ref{TTE}. 
\subsection{FuzzyPID controller}
\begin{figure}[t!]
\includegraphics[width=1.0\linewidth]{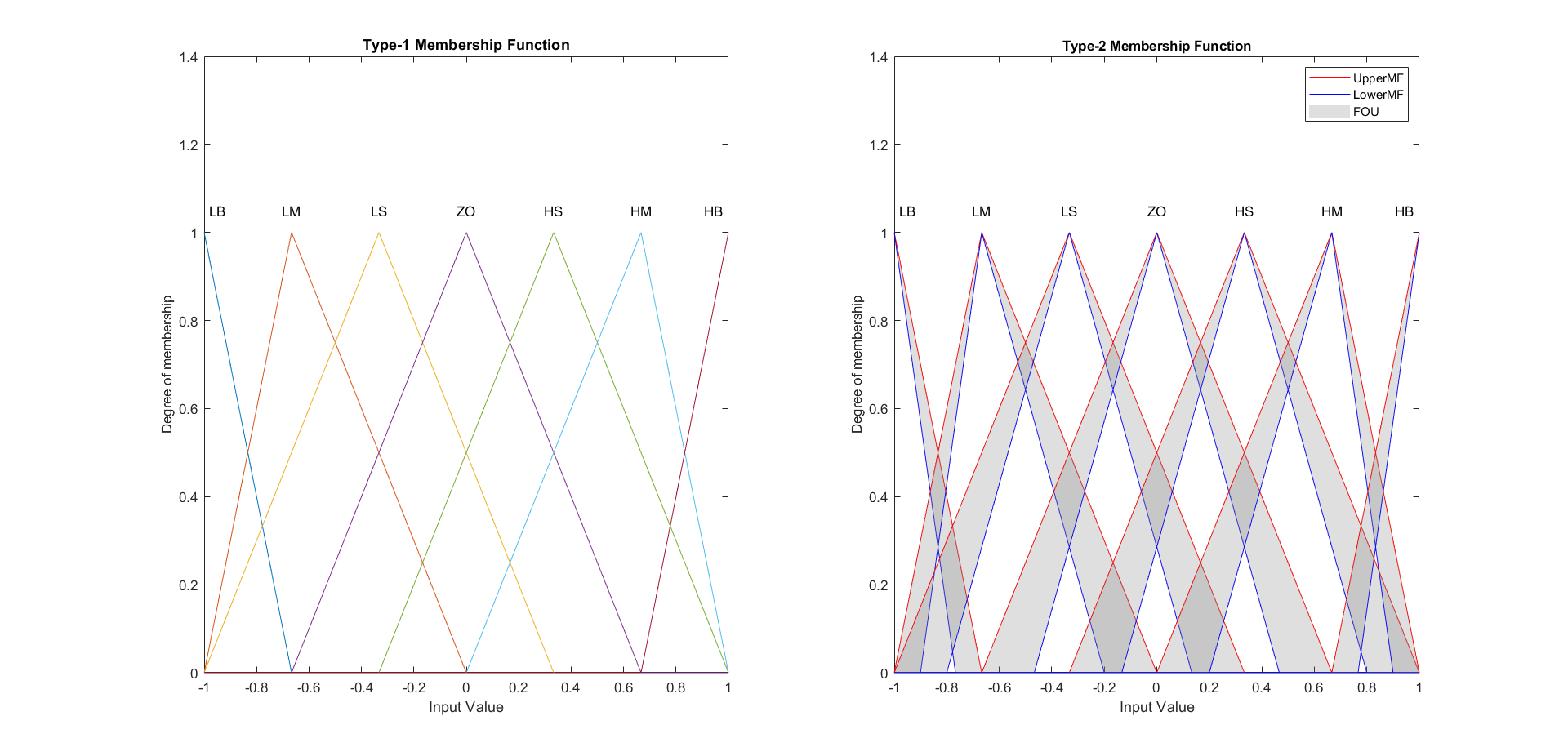}
\caption{The range of triangular membership function are adjusted based on input and output variables. Both input variables $e$ and $de$ are in the range between -1 to 1. For output variables ($kp, ki, kd$), range is adjusted in between -0.1 to 0.1 keeping the structure of membership function similar.}
\label{T_mf}
\end{figure}
The design of the type-1 fuzzy Controller begins with defining input variables $e(k)$ and the change in error $de(k)$. Initially, these are considered as crisp inputs and need to be converted into 7 overlapping fuzzy sets through the process of fuzzification. The universe of discourse for the variables is defined based on the ranges of the input variables. Triangular membership functions are employed for the fuzzification step, where the membership values (degree of truth) are obtained in the range of 0 to 1 on the universe of discourse for the given input values. This process is illustrated in Fig. \ref{T_mf}. Fuzzy sets are represented in linguistic form using the set of variables {NB, NM, NS, ZO, PS, PM, and PB}, where N, Z, H represent Negative, Zero, and Positive respectively, and B, M, S, O represent Big, Medium, Small, and Zero respectively.
\par

The next step involves designing the Mamdani-based fuzzy inference system, where a set of fuzzy rules is defined based on the relationship between fuzzy input sets ($e(k)$ and $de(k)$) and output parameters ($kp, ki, kd$). 
These rules are formulated by control engineers based on their experience with various system responses. In this context, we establish 49 rules for each T1-fuzzyPID output parameter, drawing reference from the work in \cite{xu2017speed}. A fuzzy rule takes the form of an IF(antecedent)-THEN(consequent) statement, such as "if $e(k)$ is PB and $de(k)$ is PB, then apply a large negative $kp$ (i.e., NB)". The process is executed by fuzzy inference engine that employ the max-min composition. 
The fuzzy rules for each parameter are presented in Table \ref{pid_rules}.
\begin{table}
\caption{Fuzzy rule for kp\textbackslash ki\textbackslash kd}
\begin{tabular}{|c|c|c|c|c|c|c|c|}\hline
\textbf{e}\textbackslash \textbf{de} & \textbf{NB} & \textbf{NM} & \textbf{NS} & \textbf{ZO} & \textbf{PS} & \textbf{PM} & \textbf{PB} \\
\hline \textbf{NB} & PB\textbackslash NB\textbackslash PS & PB\textbackslash NB\textbackslash NS & PM\textbackslash NM\textbackslash NB & PM\textbackslash NM\textbackslash NB & PS\textbackslash NS\textbackslash NB & ZO\textbackslash ZO\textbackslash NM & ZO\textbackslash ZO\textbackslash PS \\
\hline \textbf{NM} & PB\textbackslash NB\textbackslash PS & PB\textbackslash NB\textbackslash NS & PM\textbackslash NM\textbackslash NB & PS\textbackslash NS\textbackslash NM & PS\textbackslash NS\textbackslash NM & ZO\textbackslash ZO\textbackslash NS & NS\textbackslash ZO\textbackslash ZO \\
\hline \textbf{NS} & PM\textbackslash NB\textbackslash ZO & PM\textbackslash NM\textbackslash NM & PM\textbackslash NS\textbackslash NM & PS\textbackslash NS\textbackslash NM & ZO\textbackslash ZO\textbackslash NS & NS\textbackslash PS\textbackslash NS & NS\textbackslash PS\textbackslash ZO \\
\hline \textbf{ZO} & PM\textbackslash NM\textbackslash ZO & PM\textbackslash NM\textbackslash NS & PS\textbackslash NS\textbackslash NS & ZO\textbackslash ZO\textbackslash NS & NS\textbackslash PS\textbackslash NS & NM\textbackslash PM\textbackslash NS & NM\textbackslash PM\textbackslash ZO \\
\hline \textbf{PS} & PS\textbackslash NM\textbackslash ZO & PS\textbackslash NS\textbackslash ZO & ZO\textbackslash ZO\textbackslash ZO & NS\textbackslash PS\textbackslash ZO & NS\textbackslash PS\textbackslash ZO & NM\textbackslash PM\textbackslash ZO & NM\textbackslash PB\textbackslash ZO \\
\hline \textbf{PM} & PS\textbackslash ZO\textbackslash PB & ZO\textbackslash ZO\textbackslash NS & NS\textbackslash PS\textbackslash PS & NM\textbackslash PS\textbackslash PS & NM\textbackslash PM\textbackslash PS & NM\textbackslash PB\textbackslash PS & NB\textbackslash PB\textbackslash PB \\
\hline \textbf{PB }& ZO\textbackslash ZO\textbackslash PB & ZO\textbackslash ZO\textbackslash PM & NM\textbackslash PS\textbackslash PM & NM\textbackslash PM\textbackslash PM & NM\textbackslash PM\textbackslash PS & NB\textbackslash PB\textbackslash PS & NB\textbackslash PB\textbackslash PB \\
\hline
\end{tabular}
 \label{pid_rules}
\end{table}
\par

The final step in the design of the type-1 fuzzyPID controller is defuzzification process, which involves the conversion of linguistic variables into crisp output values using membership functions. This step is the reverse process of fuzzification, where values between 0 and 1 are converted using the Center of Gravity (COG) defuzzification technique. Ultimately, we obtain changes in the PID parameters, which are added to the previous PID parameters to continuously track the reference values, expressed as $K_{pid} = K_{pid}' + \Delta K_{pid}$. 

\par

We designed the IT2-Fuzzy PID controller for trajectory tracking in an omnidirectional robot by applying the concepts outlined in Mendel's work on uncertain systems \cite{mendel2017uncertain}. The key difference between our IT2-FPID and previous controllers lies in the use of Interval Type-2 Fuzzy Sets (IT2 FS), along with type-reduction methods that convert IT2 FS to Type-1 Fuzzy Sets (T1 FS). In this design, IT2 FS are used to represent the fuzzy input and output variables, where uncertainties are captured by the Footprint of Uncertainty (FOU), providing an additional degree of freedom. For our implementation, the error $e(k)$ and its derivative $de(k)$ are represented by seven overlapping FOUs, as illustrated in Fig. \ref{T_mf}. The upper membership functions (UMFs) of these FOUs are identical to the T1 FS, while the scaling factor for the lower membership functions (LMFs) and the lag value are set to 1.0 and 0.3, respectively. Additionally, we adopted the same rule base used in the T1 FPID controller. The firing intervals for each rule are calculated using the minimum implication method. The Karnik-Mendel (KM) algorithms are then employed to obtain T1 FSs by calculating the centroids using the iterative center of sets (COS) type-reduction method. Finally, defuzzification is performed by averaging the left and right points of the type-reduced set to produce the crisp output(PID gains).

Now, tracking the reference trajectory involves calculating the distance and direction errors from robot pose to target reference as,
\begin{equation}
    \begin{cases}
   dr=\sqrt{e_x^2 + e_y^2}, \ dr \ge threshold \\
   d\alpha=\tan^{-1}\left(\frac{y_{ref}-y}{x_{ref}-x}\right) - \theta, \ -\pi \le d\alpha \le \pi
   \end{cases}
\end{equation}
such that $e_x=x_{ref}-x$ and  $e_y=y_{ref}-y$. The direction error $d\alpha$ forces the robot to follow the target robot with linear velocity $v$ as fuzzyPID($dr$). But, this leads to problem of drifting heading angle which in our case is designed to follow the reference heading angle defined in Equation \eqref{eqn:heading error}. So, we need addition angular velocity($\omega$) to do the following task which is calculated as fuzzyPID($e_{\theta}$) such that $e_{\theta}=\theta_{ref}-\theta$. Thus, the global velocity of robot is formulated as $[v\cos(d\alpha), v\sin(d\alpha),\omega]$ which is finally passed through the inverse kinematics model to compute individual wheel velocity.
\subsection{Non-linear MPC}
The Non-linear Model Predictive Controller (NMPC) is an optimization technique that employs the non-linear classical kinematics model to predict the future behavior of a mobile robot within bounded physical constraints on both state variables and control inputs. The cost function is formulated by integrating the tracking error between the predicted states and the true output states within a finite prediction horizon $N_p$ in a sliding mode fashion. Additionally, it includes a term for minimizing the control input error for the target robot velocity. Subsequently, the cost function is processed through a non-linear solver \cite{andersson2019casadi}, thereby achieving optimal robot states and control sequences for the next $N_p$ prediction horizon. These optimal robot states are closer to the desired path based on the corresponding applied control sequences.
Next, we advance one sample time ahead to compute the target direction angle from the first state. Similarly, we consider the first control input velocity, and these steps are repeated for each sampling period with the shifted horizon.
\begin{figure}
\begin{minipage}[c]{0.45\textwidth}
  \includegraphics[width=1.05\linewidth]{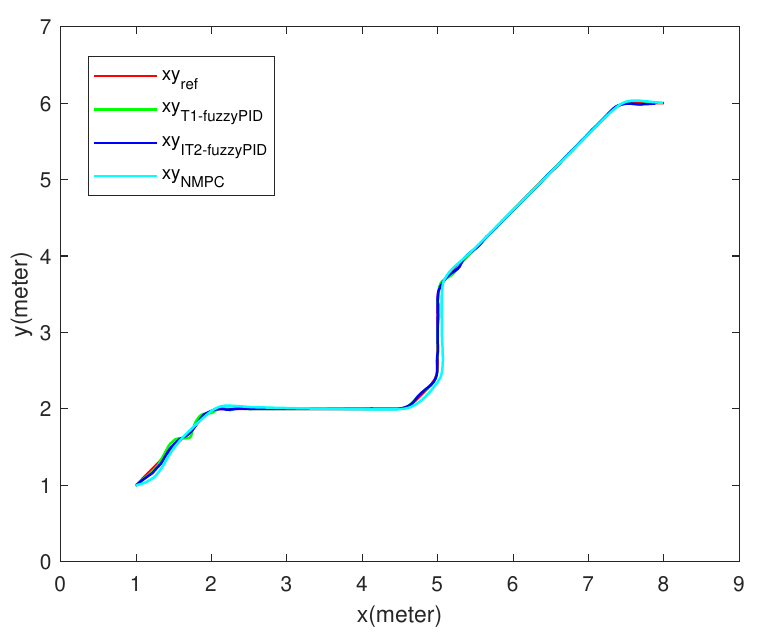}
  \caption{Comparison of tracking performance of pose(XY) for both type fuzzyPIDs and NMPC with prediction horizon of 15 with a total tracking time of 30 seconds}\label{xy_compare}
\end{minipage}%
\hspace{4 mm}
\begin{minipage}[c]{0.45\textwidth}
  \includegraphics[width=1.1\linewidth]{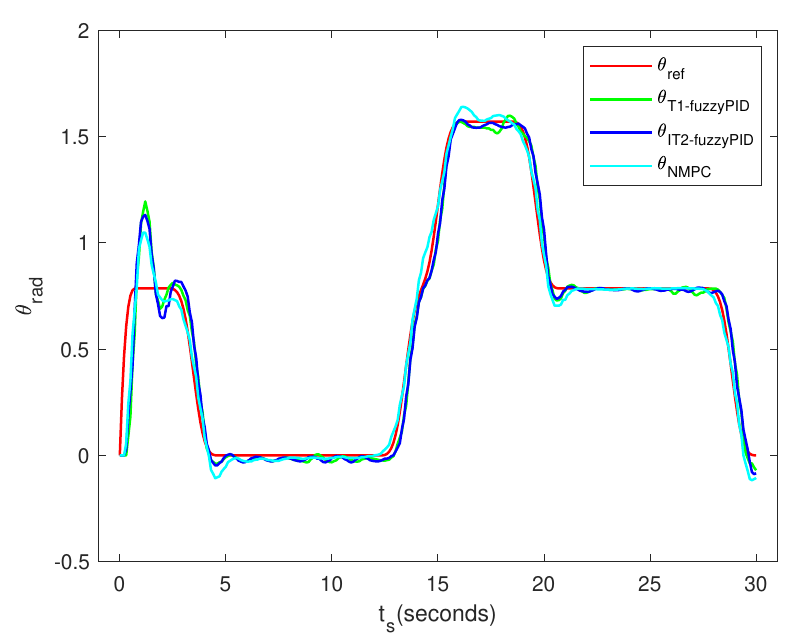}
  \caption{Comparison of tracking performance of pose($\theta$) for both type fuzzyPIDs and NMPC with prediction horizon of 15 with a total tracking time of 30 seconds}\label{theta_compare}
\end{minipage}
\end{figure}

Here, discrete time non-linear model is used to predict the evolution of future states for the mobile robot. Given $N_p$ prediction horizon, following mathematical equation is used to compute state at each sample period $k=nt_s$.
\begin{equation}
{ \mathbf{x}}(k+1 \mid k)={\mathbf{x}}(k \mid k)+t_s \dot{{\mathbf{x}}}(k \mid k)
\end{equation}
Given that,
\begin{align}
         \dot{{\mathbf{x}}}(k)&=f({\mathbf{x}}(k), {u}(k)) \ s.t. \ {\mathbf{x}}(k) \in X, {u}(k) \in U, \forall{k}\ge 0 \nonumber \\ 
         &= [v(k)cos(\theta(k)), v(k)sin(\theta(k)), \dot{\theta}(k)]^T
\end{align}
where, $\mathbf{x}(k) \in \mathbb{R}^n$ and $u(k) \in \mathbb{R}^m$ are the state and control vector respectively.
Subsequently, we calculate the error between the robot predicted states and reference trajectory as,
\begin{equation}
    \left\{\begin{array}{c}\mathbf{e}(k \mid k)=\mathbf{\Bar{x}}(k \mid k)-\mathbf{x}_{r e f}(k \mid k) \\ \vdots \\ \mathbf{e}\left(k+N_p \mid k\right)=\mathbf{\Bar{x}}\left(k+N_p \mid k\right)-\mathbf{x}_{r e f}\left(k+N_p \mid k\right)\end{array}\right.
\end{equation}

\begin{equation}
     \mathbf{x}_{r e f}(k \mid k)=\left[\begin{array}{lll}x_{r e f}(k \mid k) & y_{r e f}(k \mid k) & \theta_{r e f}(k \mid k)\end{array}\right]^{\mathrm{T}}
\end{equation}
where, $\mathbf{\Bar{x}}(k \mid k) = \mathbf{x}(k)$ which is robot initial position state. Given target velocity, we also define our objective to minimize the control input error over the finite control horizon of length $N_p-1$ as defined by equation,

\begin{equation}
    \left\{\begin{aligned}\Delta \mathbf{u}(k \mid k) & =\mathbf{\Bar{u}}(k \mid k)-\mathbf{u_{ref}}(k \mid k) \\ & \vdots \\ \Delta \mathbf{u}(k+i+1 \mid k) & =\mathbf{\Bar{u}}(k+i+1 \mid k)-\mathbf{u_{ref}}(k+i+1 \mid k) \\ & \vdots \\ \Delta \mathbf{u}\left(k+N_p-1 \mid k\right) & =\mathbf{\Bar{u}}\left(k+N_p-1 \mid k\right)-\mathbf{u_{ref}}\left(k+N_p-1 \mid k\right)\end{aligned}\right.
\end{equation}
\begin{equation}
     \mathbf{u}_{r e f}(k \mid k)=\left[\begin{array}{lll}v_{r e f}(k \mid k) & \omega_{r e f}(k \mid k) \end{array}\right]^{\mathrm{T}}
\end{equation}
Our objective cost function to be optimized is given as,
\begin{equation}
    w^*= \min J_w(\mathbf{e}(k \mid k), \Delta \mathbf{u}(k \mid k) =
    \sum_{k=0}^{N_p}\|\mathbf{e}(k+i \mid k)\|_{\mathbf{Q}}^2+\sum_{k=0}^{N_p-1}\|\Delta \mathbf{u}(k+i \mid k)\|_{\mathbf{R}}^2
\end{equation}
subject to equality constraints:
\begin{equation}
\mathbf{g}_2(\mathbf{w})=\left[\begin{array}{c}
\overline{\mathbf{x}}(k)-\mathbf{x}(k) \\
\mathbf{f}\left(\mathbf{x}(k), \mathbf{u}(k)\right)-\mathbf{x}(k+1) \\
\vdots \\
\mathbf{f}\left(\mathbf{x}(k+N_p-1), \mathbf{u}(k+N_p-1)\right)-\mathbf{x}(k+N_p)
\end{array}\right]=0
\end{equation}

where, $ w= \left[\begin{array}{lllll}\mathbf{u}_0 & \cdots & \mathbf{u}_{N_p-1}, \mathbf{x}_0 & \cdots & \mathbf{x}_{N_p}\end{array}\right]$ are problem decision variables which need to be solved. Also the weight matrix $\mathbf{Q} \in \mathbb{R}^{n*n}$ and  $\mathbf{R} \in \mathbb{R}^{m*m}$ are tuned for the better accuracy and smooth control inputs. Now, we calculate next output state $\mathbf{x_1}$ based on optimal control input $\mathbf{u_0}$. We extract third element from evolved state which acts as the target direction angle for the Omni-drive robot. Similarly, linear and angular velocity corresponding to first input vector $\mathbf{u_0}$ acts as control inputs which drives the robot to reference pose coordinates.
\\
\begin{figure}
\begin{minipage}[c]{0.45\textwidth}
  \includegraphics[width=1.1\linewidth]{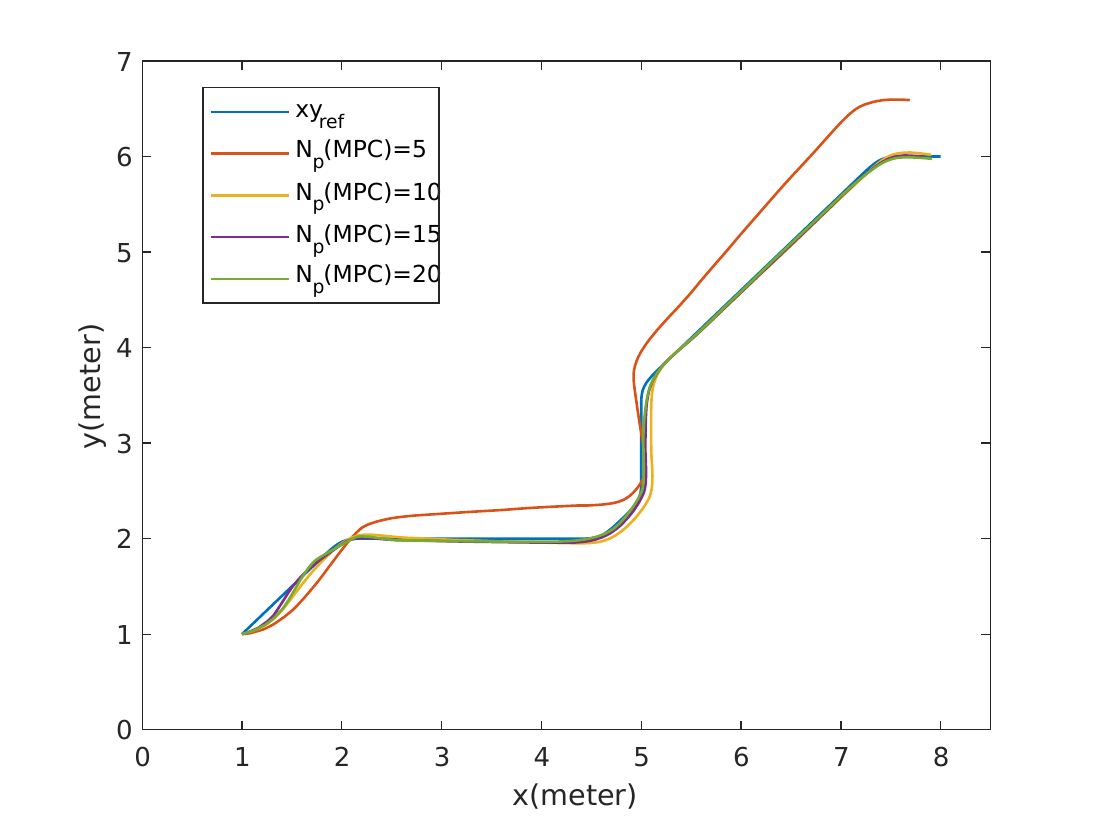}
  \caption{Comparison of tracking performance of pose(XY) for NMPC with different prediction horizon values with total tracking time of 20 seconds}\label{xy_mpc}
\end{minipage}%
\hspace{4 mm}
\begin{minipage}[c]{0.45\textwidth}
  \includegraphics[width=1.1\linewidth]{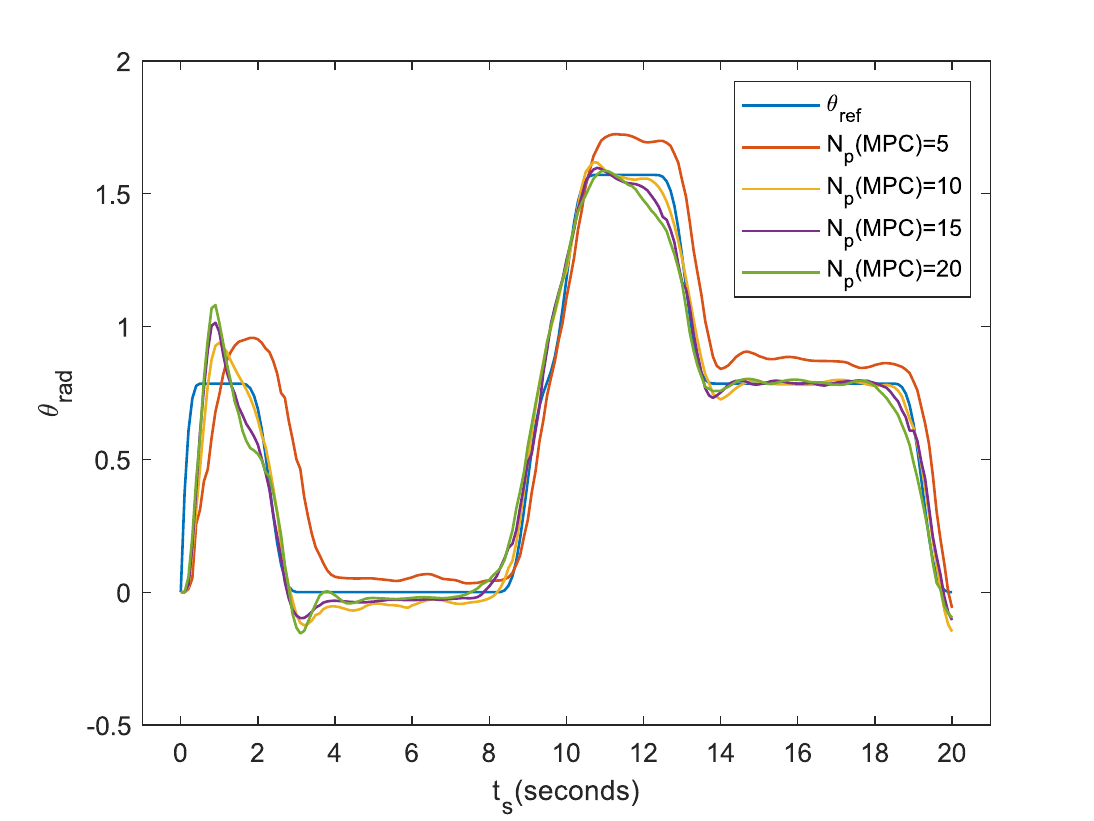}
  \caption{Comparison of tracking performance of pose($\theta$) for NMPC with different prediction horizon values with total tracking time of 20 seconds}\label{theta_mpc}
\end{minipage}
\end{figure}

\section{Simulation setup}
We conduct trajectory tracking experiments using CoppeliaSim, with our primary objective being the comparison of tracking accuracy between three controllers while keeping the sampling time constant. The tracking performance is simulated over a fixed time intervals of 20 and 30 seconds for the generated trajectory, and the results obtained are plotted for a fair comparison. We also assess the robustness of our proposed controllers in the presence of external noise. The reference and actual 2D-pose values for all the controllers are shown in Fig. \ref{xy_compare} and \ref{theta_compare}. For NMPC, we use a prediction horizon value of 15. It is crucial to note that altering the prediction horizon value has adverse effects on the controller's efficiency, as illustrated in Fig. \ref{xy_mpc} and \ref{theta_mpc}. The accuracy of the controller is also dependent on the number of intermediate poses sampled within the given tracking time. In this trajectory experiment, we choose the parameters as,
  $\mathbf{Q} = 15\mathbb{I}_3$, $\mathbf{R} = \mathbb{I}_2$, $t_s = 0.1s$, $v_{max} = 1.5 m/s$, $\omega_{max} = 3.14 rad/s$.
\par
\begin{figure}[t!]
\includegraphics[width=1.0\linewidth]{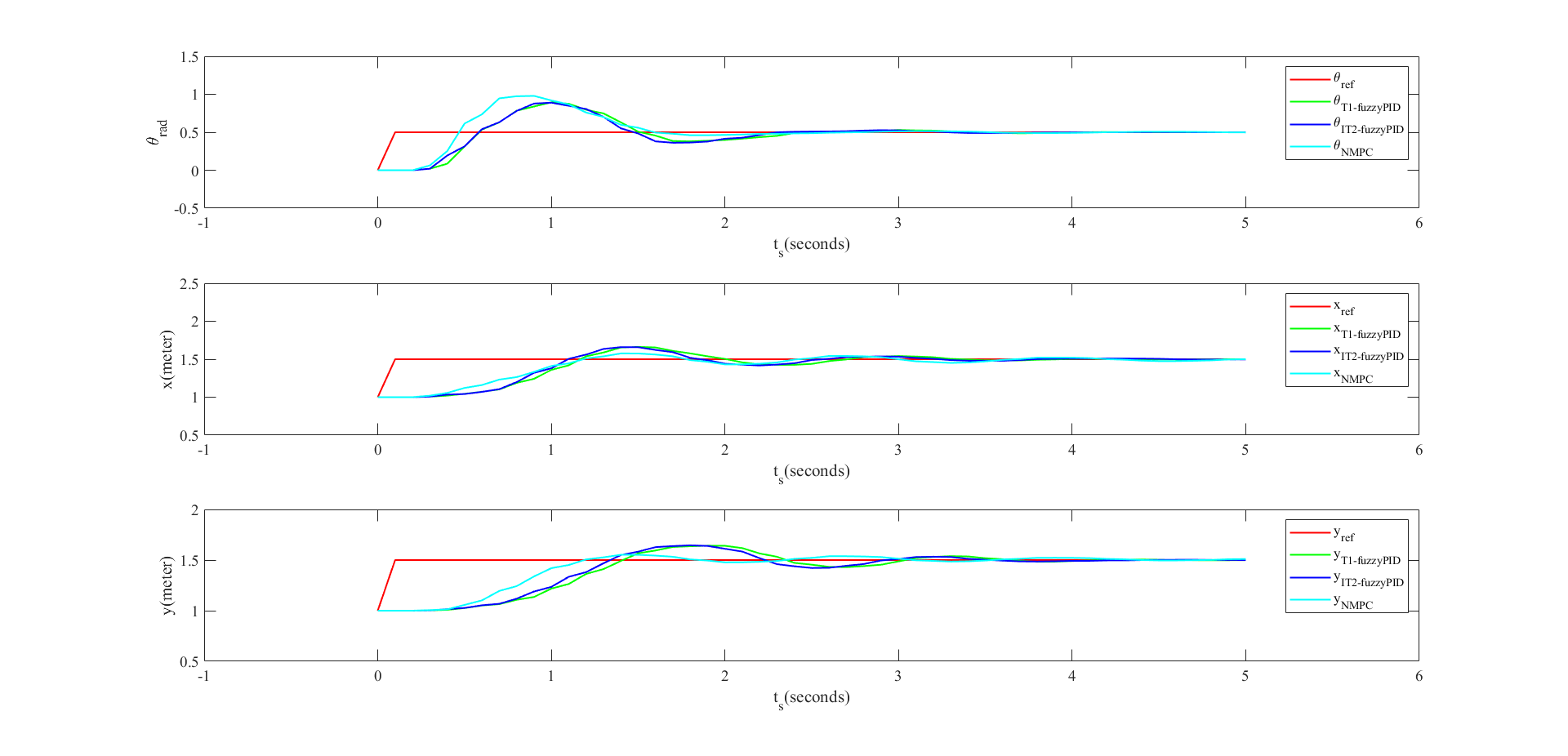}
\caption{Step response curve of our controllers}
\label{step_response_curve}
\end{figure}

We do not take hardware constraints into account as limitations for our simulation platforms. Mean Cross-track Error (ME) and Mean Absolute Error (MAE) metrics serve as the evaluation criteria for judging our results. ME is calculated by averaging the Euclidean distance between the target pose and the real pose at each sampling period. Similarly, we calculate the average value of all absolute deviations between the reference heading angle and the robot heading angle for MAE. Lower error values correspond to better tracking accuracy. Additionally, we also compute the step-response characteristics for each of our designed controllers for more detailed analysis as shown in Fig. \ref{step_response_curve}. On the other hand, we compare the tracking results for different prediction horizon values ($N_p$) in the case of Non-linear Model Predictive Control.

\begin{table}[!ht]
    \caption{Comparison of tracking error of omni drive robot for both type fuzzyPID and NMPC in absence of noise}
    \centering
\begin{tabularx}{\linewidth}{ c *{12}{c} }
    \toprule
     & \multicolumn{2}{c}{\textbf{T1-fuzzyPID}}
    &  \multicolumn{2}{c}{\textbf{IT2-fuzzyPID}}
            & \multicolumn{2}{c}{\textbf{NPMC($N_p=15$)}}   \\
     \toprule
 \textbf{Tracking time}   &   \textbf{ME($XY$)}  &   \textbf{MAE($\theta$)}  &   \textbf{ME($XY$)}  &   \textbf{MAE($\theta$)}  & \textbf{ME($XY$)}  &   \textbf{MAE($\theta$)} \\
    \midrule
 20 seconds   &   0.0874 m &  0.0888 rad  &  0.0855 m & 0.0849 rad &   \textbf{0.0722 m} & \textbf{0.0625 rad} \\
 30 seconds   &   0.0521 m & 0.05349 rad   &   0.0501 m & 0.0535 rad &   \textbf{0.0439 m} & \textbf{0.0385 rad} \\
    \bottomrule
\end{tabularx}
 \label{error_compare}
\end{table}
\section{Comparison of fuzzyPIDs and NMPC}
\subsection{In Absence of Noise}
Table \ref{error_compare} presents the tracking error for both fuzzyPIDs and NMPC with a prediction horizon value of 15 in the absence of noise. The NMPC controller achieves a lower error compared to fuzzyPIDs in both metrics while a similar tracking performance is evaluated for fuzzy based controllers. As the tracking time increases, a corresponding improvement in tracking accuracy is observed for both controllers. On the other hand, we found fuzzy Controllers to be more computationally efficient than NMPC, albeit at the expense of tracking accuracy.

\begin{figure}[t!]
\includegraphics[width=0.7\linewidth]{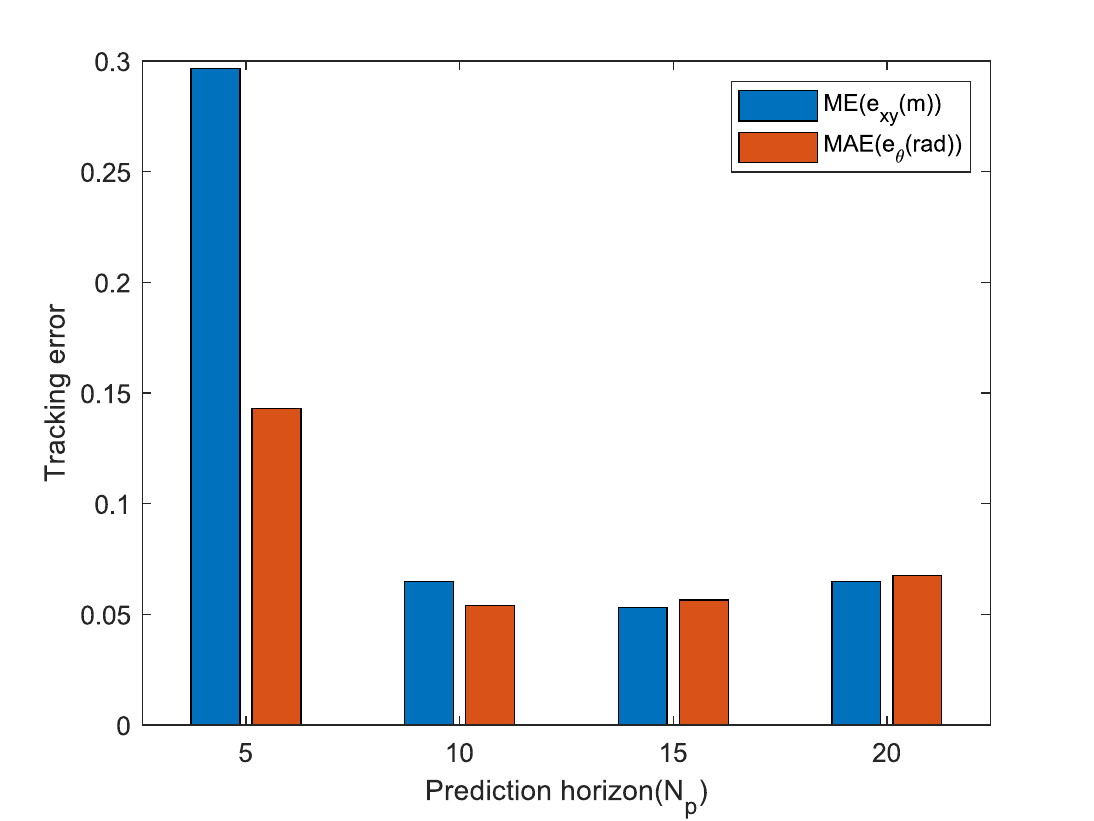}
\caption{Comparison of tracking error of robot pose for NMPC with different prediction horizon values with total tracking time of 20 seconds}
\label{error_bar_mpc}
\end{figure}

Conversely, the performance of NMPC is contingent upon selecting a suitable value for the prediction horizon, resulting in a significant impact on the designed controller's performance. Opting for a lower prediction horizon value leads to suboptimal control actions, rendering the controller less adept at handling the complex dynamics of the system and resulting in lower accuracy. Conversely, a larger prediction horizon increases system complexity and results in a slower response. Furthermore, it introduces a higher risk of overfitting to the prediction model, subsequently yielding sub-optimal results. The error for different values of the prediction horizon is illustrated in Fig. \ref{error_bar_mpc}.

\begin{table}[!ht]
    \caption{Comparison of tracking error of omni drive robot for both type fuzzyPIDs and NMPC in presence of noise}
    \centering
\begin{tabularx}{\linewidth}{ c *{12}{c} }
    \toprule
     & \multicolumn{2}{c}{\textbf{T1-fuzzyPID}}
    &  \multicolumn{2}{c}{\textbf{IT2-fuzzyPID}}
            & \multicolumn{2}{c}{\textbf{NPMC($N_p=15$)}}   \\
     \toprule
 \textbf{Tracking time}   &   \textbf{ME($XY$)}  &   \textbf{MAE($\theta$)}  &   \textbf{ME($XY$)}  &   \textbf{MAE($\theta$)}  & \textbf{ME($XY$)}  &   \textbf{MAE($\theta$)} \\
    \midrule
 30 seconds   &   0.0657 m & 0.0608 rad   &   0.0647 m & 0.0588 rad &   \textbf{0.0566 m} & \textbf{0.0586 rad} \\
    \bottomrule
\end{tabularx}
 \label{noise_error_compare}
\end{table}

\subsection{Random Noise Injection}

\begin{figure}
\begin{minipage}[c]{0.45\textwidth}
  \includegraphics[width=1.05\linewidth]{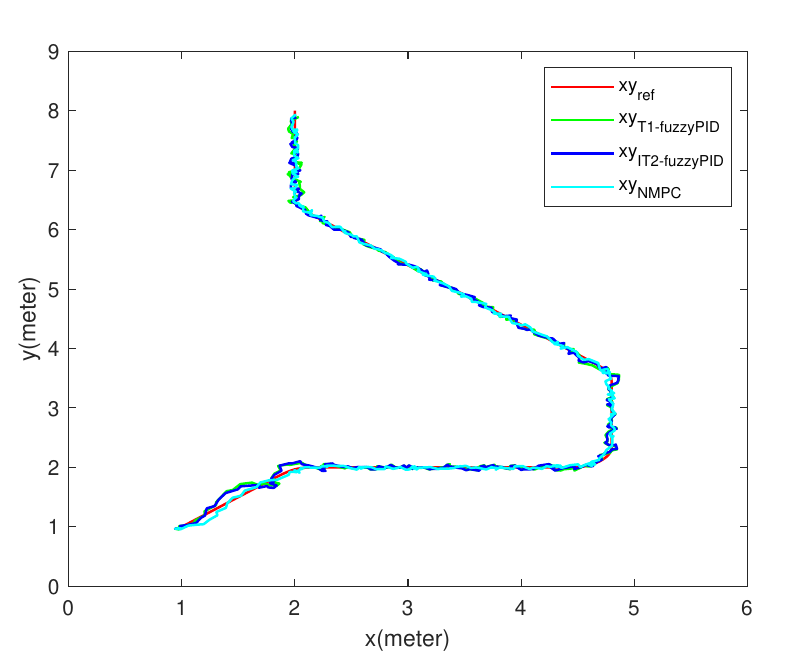}
  \caption{Comparison of noise rejection capabilities of our controllers on pose(XY)}\label{noisy_xy_plot}
\end{minipage}%
\hspace{4 mm}
\begin{minipage}[c]{0.45\textwidth}
  \includegraphics[width=1.1\linewidth]{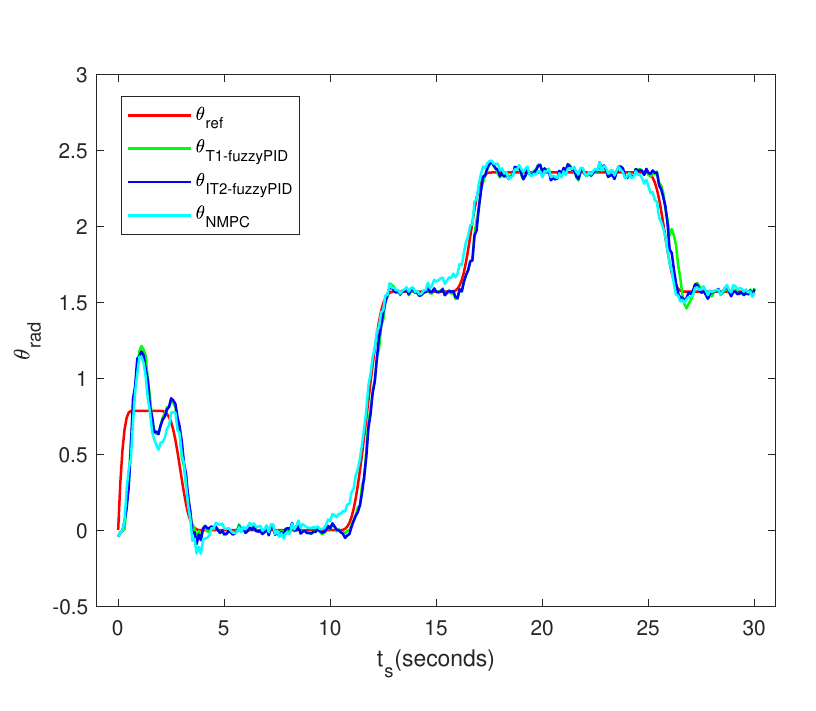}
  \caption{Comparison of noise rejection capabilities of our controllers on pose($\theta$)}\label{noisy_theta_plot}
\end{minipage}
\end{figure}

To assess the noise rejection capabilities of our proposed controllers for the Omni-drive robot, we introduced randomly generated noise into the feedback path of the input pose state at each timestamp, modeled by the equation $\dfrac{rand()}{6} \times \sin(\dfrac{nt_s}{5})$. Simulations were conducted over 30 seconds, with the resulting trajectory responses displayed in Fig. \ref{noisy_xy_plot} and Fig. \ref{noisy_theta_plot}. The NMPC controller is found to be more robust to external noise compared to all other controllers. Additionally, the IT2-fuzzyPID controller exhibited greater noise rejection capability than the T1-fuzzyPID controller, attributable to its ability to model uncertainties, despite showing similar tracking performance in the absence of noise. The Mean Error (ME) and Mean Absolute Error (MAE) metrics for the proposed controllers in presence of noise are summarized in Table \ref{noise_error_compare}.

\begin{table}[!ht]
    \centering
 \resizebox{\textwidth}{!}{
\begin{tabular}{|l|ccc|ccc|ccc|}
        \hline
        \textbf{Metric} & \multicolumn{3}{c|}{\textbf{T1-fuzzyPID}} & \multicolumn{3}{c|}{\textbf{IT2-fuzzyPID}} & \multicolumn{3}{c|}{\textbf{NMPC}} \\
        \hline
        & \( x \) & \( y \) & \( \theta \) & \( x \) & \( y \) & \( \theta \) & \( x \) & \( y \) & \( \theta \) \\
        \hline
        \( M_{\mathrm{p}} \% \) & 10.800 & 9.448 & 78.904 & 10.611 & 9.692 & \textbf{77.887} & \textbf{5.083} & \textbf{3.480} & 95.766 \\
        \( t_r, 0.1 \rightarrow 0.9 \) & 0.991 & 1.186 & 0.215 & \textbf{0.949} & 1.133 & 0.242 & 2.252 & \textbf{0.914} & \textbf{0.175} \\
        \( t_s \pm 10\% \) & 1.610 & 1.186 & 2.282 & 1.522 & 1.133 & 2.160 & \textbf{0.921} & \textbf{0.914} & \textbf{1.514} \\
        \hline
    \end{tabular}
    }
\caption{Step-response characteristics, such as overshoot($M_{\mathrm{p}} \%$), rise time($\boldsymbol{t}_r$) and settling time($\boldsymbol{t}_s$) of our proposed controllers}
\label{step_response_table}
\end{table}

\subsection{Step Response Analysis}
 The step-response analysis reveals that the NMPC controller achieves the fastest response, characterized by the minimal overshoot, shortest rise time and settling time, thereby ensuring superior stability. This performance is followed closely by the IT2-fuzzyPID controller, which outperforms the T1-fuzzyPID controller in trajectory tracking. Detailed insights into the step-response characteristics of all controllers are provided in Table \ref{step_response_table}.

\section{Conclusions}
In this study, we propose a comprehensive framework for developing a trajectory tracking methodology in a 4-wheel Omni-drive robot. This scheme involves deriving the kinematic model, creating an obstacle-free path, followed by a path-smoothing algorithm. Subsequently, we design two popular control methods, namely fuzzy based PIDs and NPMC, for achieving robust path tracking. The entire setup is simulated, tracking performance is compared by calculating the cross-track error for all of our controllers. Additionally, step-response parameters are computed for clearly highlighting the gap seen in each controllers.
\par
The results indicate that NMPC outperforms fuzzyPIDs, albeit with a trade-off in computational complexity. The NMPC has the better step-response characteristics with low tracking error rate and higher noise rejection capability. Conversely, the simplicity of fuzzyPIDs makes it applicable in a wide variety of controller applications. We also assess the effectiveness of NMPC for different values of the prediction horizon in the stability control of our robot. 
\par
In a nutshell, the proposed approaches can be useful for a variety of control and robotics tasks, aiming to improve the accuracy and throughput of the system.

\section{Limitations and Future Enhancements}
While the performance of the aforementioned controllers has been evaluated in simulations, real-time implementation remains untested as they are bound to various external disturbances influencing the stability of mobile robots. Additionally, the simulations were conducted in a static environment limiting their use-case on real world scenarios. Therefore, future work will focus on the real-time implementation of these controllers in both static and dynamic environments to better understand their practical efficacy and adaptability. We also aim to compare the tracking performance of newly proposed IT3 fuzzy system with our proposed approach. Moreover, We also aimed to tune the the sensible gain parameters with the use of various evolutionary algorithms as mentioned in various literature for future work.


\bibliography{revised_sn-bibliography}

\end{document}